\documentclass{article}
\usepackage{spconf,amsmath,graphicx,hyperref}
\usepackage{booktabs}
\usepackage{makecell}
\usepackage{amssymb}
\usepackage{CJKutf8}
\makeatletter
\renewcommand\section{\@startsection{section}{1}{\z@}{2.1ex plus .6ex minus .2ex}{1.2ex plus .1ex}{\normalfont\Large\bfseries}}
\renewcommand\subsection{\@startsection{subsection}{2}{\z@}{1.7ex plus .5ex minus .2ex}{0.8ex plus .1ex}{\normalfont\large\bfseries}}
\renewcommand\subsubsection{\@startsection{subsubsection}{3}{\z@}{1.5ex plus .4ex minus .2ex}{0.7ex plus .1ex}{\normalfont\normalsize\bfseries}}
\makeatother

\usepackage[dvipsnames]{xcolor}

\newcommand{\ja}[1]{\begin{CJK}{UTF8}{min}#1\end{CJK}}

\title{WHEN THE PERSONA CONTRADICTS THE LINE: \\DIAGNOSING PLAN-TO-SYNTHESIS FAILURES IN CHAIN-OF-THOUGHT TTS}

\name{Joonyong Park, Jerry Li}
\address{Spellbrush, CA, USA\\
\texttt{\small jyjoon97@gmail.com, jerry@sizigistudios.com}}

\begin{document}
\ninept
\maketitle

\begin{abstract}
Chain-of-thought TTS systems rely on language-model planners to translate persona and context into vocal delivery, but it is unclear where this intent is lost when the requested affect conflicts with lexical sentiment.
We introduce a bilingual Japanese--English persona--line conflict benchmark in which each item pairs one line with congruent and conflicting personas and pre-declares the expected acoustic direction.
We probe this direction in the planner-generated delivery phrase, the planner-written emotion vector, the QwenEmotion-generated vector, and the rendered waveform.
Planner outputs are usually direction-correct, while downstream agreement can be substantially lower.
Bypassing QwenEmotion provides only modest gains over the full planner-controlled Base.
Supervising QwenEmotion raises the direction rate to 75\% in Japanese and 84\% in English.
A learned text-to-latent interface reaches the highest English rate of 89\% and the highest Japanese arousal rate of 88\%.
These results show that learned repair of the plan-to-synthesis interface can recover counter-textual delivery, although its effect differs across languages and acoustic axes.
\end{abstract}

\begin{keywords}
expressive TTS, chain-of-thought, persona TTS, emotion recognition
\end{keywords}

\section{Introduction}
\label{sec:intro}
Instruction-conditioned text-to-speech (TTS) adjusts its delivery to natural-language descriptions of speaker, situation or tone \cite{du2024cosyvoice2,zhou2026flexivoice}. 
To achieve this, recent systems add a planning step in which a language model scripts how a line should be spoken before any audio is generated \cite{ling2025batonvoice}.
When the requested affect agrees with the words, such systems follow instructions with high perceived quality \cite{indextts2,du2024cosyvoice2}.

However, dramatic acting frequently requires the exact opposite: a character might offer congratulations while deeply grieving, or speak cheerful words while utterly exhausted.
We refer to this as a \emph{persona--line conflict}, where the required vocal delivery must contradict the underlying sentiment of the written text.
In these cases, current instruction-following synthesizers typically fail, defaulting to the text's inherent emotion far more often than they follow the prompt \cite{zhou2026flexivoice,chen2026mint}.
Recent evaluations acknowledge this failure \cite{zhou2026flexivoice}, but they rely on broad categorical labels or overall ratings.
Because they only judge the final audio output, they cannot pinpoint where the system breaks down: did the planner choose the wrong delivery, did the emotion-control interface fail to preserve it, or did the synthesizer fail to realize the intended acoustics?
Without probing these intermediate representations, researchers cannot determine which component should be modified, and a reported improvement may be difficult to localize.

To answer this question, we introduce a test set in Japanese and English that isolates persona--line conflicts.
Each item pairs one spoken line with a congruent persona, under which the line reads naturally, and a conflicting persona that requires an opposing affect.
The item also declares an acoustic target $(a,s)$, specifying whether arousal or valence should increase or decrease between the two renditions.
Because this target is determined before synthesis, the same declared direction can be used to evaluate both intermediate representations and the rendered speech.
Declaring an acoustic axis also avoids requiring each persona to correspond to a single emotion category.
We score the generated speech with a predictor read on the declared axis \cite{wagner2023dawn} and validate its pairwise direction judgments on parallel human recordings \cite{zhou2021esd,adigwe2018emotional}.

We apply this protocol to a planner-controlled IndexTTS-2.5 pipeline.
Given the persona and spoken line, the external planner produces a free-form delivery phrase and an eight-dimensional emotion vector.
In the stock route, QwenEmotion~\cite{huang2025emotionqwen} independently converts the delivery phrase into another eight-dimensional emotion vector, which is mapped through a speaker-matched emotion prototype bank to the conditioning representation used for synthesis.
We probe agreement with the declared direction at four points: the delivery phrase, the planner-written emotion vector, the QwenEmotion-generated emotion vector, and the waveform produced by the stock emotion route with planner-side duration control disabled.

We then evaluate three modifications of this emotion-control interface while keeping the synthesis backbone fixed.
First, we bypass QwenEmotion and pass the planner-written emotion vector directly to the existing prototype bank without additional training.
Second, we fine-tune QwenEmotion using render-selected supervision.
Third, we replace both QwenEmotion and the prototype bank with a learned text-to-latent network that predicts a continuous emotion-conditioning shift directly from the planner-generated delivery phrase.

Our contributions are (i) a bilingual persona--line conflict benchmark with a pre-declared acoustic target and reference-free scoring protocol; (ii) a representation-level diagnosis that probes direction agreement in the delivery phrase, planner-written emotion vector, QwenEmotion-generated emotion vector, and rendered waveform; and (iii) a controlled comparison of bypassed, supervised, and learned emotion-control interfaces while keeping the synthesis backbone fixed.



\begin{table*}[t]
\centering
\caption{Example test items. The target is fixed before any audio exists.}
\label{tab:examples}
\footnotesize
\setlength{\tabcolsep}{5pt}
\begin{tabular}{p{4.4cm}p{5.4cm}p{5.8cm}c}
\toprule
Line & $P_{\mathrm{cong}}$ & $P_{\mathrm{conf}}$ & $(a,s)$ \\
\midrule
``\ja{美味しいです}'' (It's delicious.) & a child praising the meal a parent cooked & a wedding guest eating food they cannot enjoy & val $-$ \\
``You missed the tram by about ten seconds.'' & a helpful stranger on a rainy platform, to a commuter who arrived too late & a weary parent in the dark kitchen, as the exhausted teenager comes home long after curfew & ar $-$ \\
\bottomrule
\end{tabular}
\end{table*}

\textbf{Related work.}
Benchmarks of instruction-following TTS do not declare an acoustic target in advance.
As a result, they cannot locate a breakdown between the plan, the control parameters and the waveform \cite{ling2025batonvoice,zhou2026flexivoice,chen2026mint}.
Speech emotion recognizers can rely on the sentiment of the spoken words \cite{wagner2023dawn,triantafyllopoulos2022probing}.
This lexical bias is problematic for evaluating speech whose intended affect conflicts with the text.
Parallel recordings of the same sentence under different emotions have been used in studies of mismatched perceptual channels \cite{mitchell2006incongruent} and in constructing preference pairs \cite{zhou2026flexivoice}.
Such recordings have not been used to validate a metric for counter-textual direction \cite{zhou2021esd,adigwe2018emotional}.
Natural-language style prompts can be mapped to style embeddings by prompt encoders trained jointly with the synthesizer \cite{guo2023prompttts,yang2024instructtts}.
Other work aligns text and speech style in a shared representation space \cite{liu2023promptstyle}.
Emotion can also be controlled through categorical vectors and reference or exemplar embeddings \cite{indextts2,cho2024emosphere}.
Our learned interface instead predicts the pretrained synthesizer's own emotion latent from text using targets extracted from human acted speech.
The synthesizer itself remains fixed.

\begin{figure*}[t]
\centering
\includegraphics[width=0.99\textwidth]{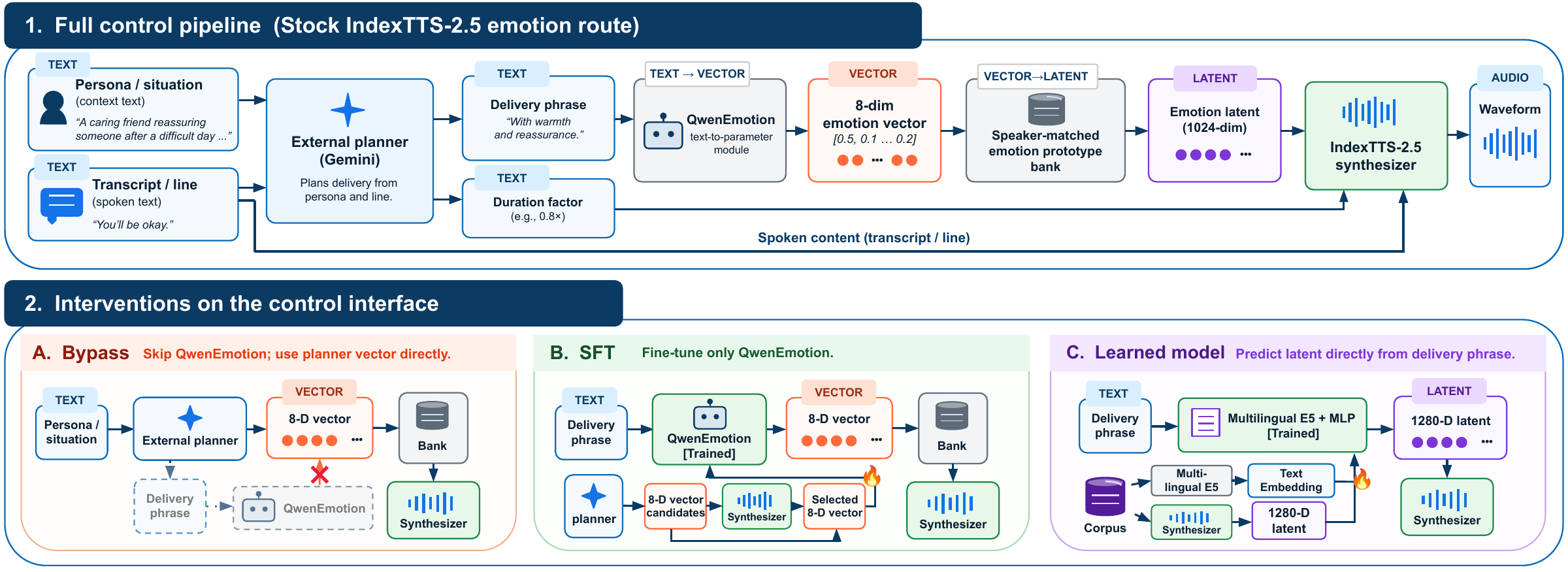}
\caption{Stock IndexTTS-2.5 emotion-control route and the three interface interventions evaluated in this work.}
\label{fig:interface_routes}
\end{figure*}

\vspace{-2mm} 
\section{Method}
\label{sec:method}

\subsection{Conflict benchmark and scorer}
\label{ssec:bench}

Each item comprises a line $\ell$, a congruent persona $P_{\mathrm{cong}}$, a conflicting persona $P_{\mathrm{conf}}$, and a declared target $(a,s)$. 
The target specifies an axis $a \in \{\text{arousal}, \text{valence}\}$ and a direction $s \in \{+,-\}$.
$P_{\mathrm{cong}}$ elicits a natural reading of the line, whereas $P_{\mathrm{conf}}$ places the same words in a context requiring an opposing affect.
A conflicting rendition is direction-correct when it moves along $a$ in direction $s$ relative to the congruent rendition.

An LLM authors each item in two steps, first proposing a scene and then introducing a conflicting twist.
A planner-independent jury retains an item only when at least four of five seeded votes agree with its declared direction.
A fresh five-vote audit of 75 items reproduced the declared target on 96\% of items.
The authors then reviewed the retained items and vetoed ambiguous cases.
Each language contains 150 test items, with 56 arousal and 94 valence items.
Japanese speech is rendered with Tsukuyomi-chan\footnote{\url{https://tyc.rei-yumesaki.net/material/corpus/}} and Amitaro\footnote{\url{https://amitaro.net/voice/corpus-list/ita/}}, and English speech with the EmoV-DB speakers Bea and Sam \cite{adigwe2018emotional}.
All training lines and training voices used in the experiments are disjoint from the test set.

We score rendered speech with a wav2vec2-based acoustic predictor, which maps an utterance to continuous arousal, dominance and valence in [0, 1] from the audio alone\cite{wagner2023dawn}. 
The regressor is trained on dimensional arousal, dominance and valence annotations from MSP-Podcast \cite{busso2026msppodcast}.
Let $m_a(\cdot)$ denote the output of a regressor on the declared axis $a$.
An item is direction-correct when the sign of $\Delta = m_a(x_{\mathrm{conf}}) - m_a(x_{\mathrm{cong}})$ equals the declared direction $s$.

Because the regressor is trained only on MSP-Podcast, we validate it on out-of-domain parallel human recordings.
We use same-speaker, same-sentence recordings in different emotional states from Voice-Statistics\footnote{\url{https://voice-statistics.github.io/}}, the Amitaro ITA recordings, EmoV-DB \cite{adigwe2018emotional}, and ESD \cite{zhou2021esd}.
Across 24{,}334 arousal pairs, the regressor recovers the labelled direction on 99.6\%; across 15{,}601 valence pairs, the corresponding rate is 83.1\%.

The direction rate for each condition is the per-item majority over three plan seeds.
We compare conditions with two-sided exact sign tests over discordant items.
All reported renders satisfy a Whisper-derived CER \cite{radford2023whisper} of at most 0.15 for Japanese and 0.25 for English.
UTMOS \cite{saeki2022utmos} remains within 0.14 of the base in Japanese and above the base in English.

\subsection{Pipeline and representation probe}
\label{ssec:pipeline}

Fig.~\ref{fig:interface_routes} explains the whole pipeline. 
We use IndexTTS-2.5 \cite{indextts2} as a fixed synthesis backbone and place an external LLM planner \texttt{Gemini~3.5 Flash-Lite} in front of it. 
For each persona--line pair, the planner produces a free-form delivery phrase, an eight-dimensional emotion vector, a scalar strength $\alpha$, and a duration factor.
The spoken line is passed separately to the synthesizer model as the transcript.
In the stock emotion route, the delivery phrase is converted by QwenEmotion\cite{huang2025emotionqwen} into an eight-dimensional emotion vector, which is mapped through the speaker-matched prototype bank to the emotion-conditioning representation used for synthesis.
The planner-predicted duration factor is supplied directly to the synthesizer and is shared across the main interface conditions.

We probe agreement with the declared direction at four points in the emotion-control path.
The delivery phrase is scored by the planner-independent jury used for benchmark construction.
The planner-written and QwenEmotion-generated emotion vectors are projected onto the declared axis using a fixed sign map over the eight emotion categories.\footnote{Arousal high: happy, angry, afraid, surprised; low: sad, melancholic, calm. Valence positive: happy, calm; negative: angry, sad, afraid, disgusted, melancholic. Unlisted categories receive zero weight on that axis.}
The waveform is scored by the acoustic contrast $\Delta$ defined in Sec.~\ref{ssec:bench}.
The duration factor is not included in this representation probe because it is a separate prosodic control supplied directly to the synthesizer.
Accordingly, for the waveform probe, we fix the duration factor to 1.0 to isolate the stock emotion-control path.

\vspace{-1mm} 

\subsection{Interface treatments}
\label{ssec:treat}

All treatments keep the IndexTTS-2.5 synthesis backbone fixed and differ only in how the planner output is converted into emotion conditioning.
For each persona condition, the planner predicts a duration factor in $[0.7,1.4]$ from the situation and delivery phrase.
Unless otherwise stated, all main conditions use this planner-predicted duration factor. 
In the \emph{w/o duration} ablation, the same emotion-control route is used while the duration factor is fixed to 1.0.
In the \emph{w/o conditioning} ablation, we replace the planner-dependent emotion conditioning with the neutral conditioning used by the synthesis backbone while retaining the planner-predicted duration factor.
The \emph{w/o duration \& conditioning} ablation removes both controls.

\textbf{Bypassing QwenEmotion.}
The structured route bypasses QwenEmotion and passes the planner-written eight-dimensional emotion vector and scalar strength $\alpha$ directly to the same speaker-matched prototype bank.
The prototype bank, planner-side duration control, and synthesis backbone remain unchanged.

\textbf{Supervising QwenEmotion.}
We fine-tune only QwenEmotion, whose input is the planner-generated delivery phrase and whose output is the eight-dimensional emotion vector used by the stock route.
We initially construct 800 training items per language using \texttt{Gemini~3.5 Flash-Lite}, each specifying a line, situation, delivery phrase, and declared affective axis and direction.
Training voices are disjoint from the test voices: three Voice-Statistics speakers and one JVNV speaker for Japanese, and two EmoV-DB speakers for English.

For each item, we render one neutral reference and five candidate $(\mathbf{v},\alpha)$ controls with the synthesis backbone fixed.
Candidate vectors are constructed from emotion categories consistent with the declared direction. 
Each candidate is evaluated by its acoustic contrast against the neutral rendition on the declared axis, subject to CER $\leq 0.15$ for Japanese and $\leq 0.25$ for English.
We additionally reject candidates that induce an excessive off-axis change inconsistent with the delivery phrase's LLM-labelled valence, or whose dominant emotion category contradicts that valence.

Among the remaining candidates, we select the lowest-$\alpha$ candidate whose contrast exceeds the pilot threshold of 0.15 for arousal or 0.30 for valence.
If no candidate exceeds the threshold, we select the candidate with the largest contrast in the declared direction; items for which no candidate moves in the required direction are discarded.
This procedure yields 569 Japanese and 524 English training pairs, of which 293 and 342, respectively, exceed the corresponding threshold.
The supervision target is the selected emotion vector scaled by $\alpha$ in QwenEmotion's native output format, since the stock text route does not expose a separate strength output.
We fine-tune QwenEmotion with LoRA \cite{hu2022lora} using rank 16, scaling 32, dropout 0.05, four epochs, a learning rate of $10^{-4}$, and batch size 8.

\textbf{Learning a text-to-latent interface.}
We train a model that maps a natural-language delivery description directly to a continuous 1{,}280-dimensional emotion-conditioning representation.
The training targets are constructed from 30{,}083 acted clips of 27 speakers disjoint from the test voices: 14{,}345 English clips (ESD \cite{zhou2021esd}, 9 speakers; EmoV-DB \cite{adigwe2018emotional}, 2 speakers), 14{,}000 Mandarin clips (ESD, 10 speakers), and 1{,}738 Japanese clips (JVNV \cite{xin2024jvnv}, 3 speakers; Voice-Statistics, 3 speakers).

For each emotional recording, we pair it with the same speaker's neutral recording of the same sentence and pass both through the IndexTTS-2.5 audio-prompt pathway.
Let $A(x)\in\mathbb{R}^{1280}$ denote the extracted emotion-conditioning representation.
For an emotional recording $x_i^{\mathrm{emo}}$ and its matched neutral recording $x_i^{\mathrm{neu}}$, we define the target shift as
\begin{equation}
\Delta\mathbf{z}_i
=
A(x_i^{\mathrm{emo}})
-
A(x_i^{\mathrm{neu}}).
\end{equation}
Because the paired recordings share the speaker and linguistic content, this difference is used as an audio-derived target for the change associated with emotional delivery.

Each target shift is paired with a free-form delivery description $d_i$ generated from the source corpus's emotion annotation using \texttt{Gemini~3.5 Flash-Lite}.
The description is generated without access to the recording or its extracted latent representation.
A frozen multilingual E5-large encoder \cite{wang2024multilinguale5} maps $d_i$ to a 1{,}024-dimensional text embedding, and a two-layer MLP $f_\theta$ predicts the corresponding 1{,}280-dimensional emotion-conditioning shift:
\begin{equation}
\hat{\Delta\mathbf{z}}_i
=
f_\theta\!\left(E(d_i)\right),
\qquad
E(d_i)\in\mathbb{R}^{1024},
\quad
\hat{\Delta\mathbf{z}}_i\in\mathbb{R}^{1280}.
\end{equation}
The MLP is trained to match the audio-derived target shift using mean-squared error and cosine loss, while the E5 encoder and synthesis backbone remain fixed.

At inference, the planner-generated delivery phrase is encoded by E5 and mapped by the trained MLP directly to an emotion-conditioning shift.
The predicted shift is added to the neutral emotion-conditioning representation of the target speaker to obtain the conditioning representation used for synthesis.
This bypasses both QwenEmotion's category representation and the subsequent prototype-bank mapping.

\begin{table}[t]
\centering
\caption{Agreement with the declared direction at four probe points over 150 items per language.
Each rate is the per-item majority over three plan seeds.
Phrase--vector rows report item counts.}
\label{tab:stages}
\footnotesize
\begin{tabular}{lcc}
\toprule
Probe point & JP & EN \\
\midrule
Delivery phrase (LLM jury) & 84\% & 98\% \\
Planner-written emotion vector (sign map) & 86\% & 89\% \\
\quad agreement with delivery phrase & 83\% & 89\% \\
\quad phrase-only / vector-only correct (items) & 11 / 14 & 15 / 1 \\
QwenEmotion vector (sign map) & 46\% & 83\% \\
\quad arousal / valence & 61 / 37\% & 82 / 84\% \\
Waveform (stock, w/o duration) & 55\% & 66\% \\
\bottomrule
\end{tabular}
\end{table}

\textbf{Listening test.}
Ten listeners per language perform a blinded matching test on 30 items drawn at random from the test set, each rendered under three conditions of Table~\ref{tab:scoreboard}: the base, the supervised module and the learned interface, all with the planner's duration.
In every trial the listener hears the two renditions of one item and assigns each to its situation; the condition is hidden, and trial order and the positions of audio and text are randomised per listener.
The 90 trials per language are split into two lists of 45, each taken by five listeners, so every item receives five judgments per condition.
An item counts as correct for a condition when at least three of its five judgments are. 
Conditions are tested against chance by two-sided exact sign tests over items, and we report the item-level agreement between the listener majority and the predictor's verdict on the served renditions.

\section{Experiments and Results}
\label{sec:results}

\subsection{Where direction agreement breaks down}
\label{ssec:res-decomp}

We probe the declared direction in the planner-generated delivery phrase, the planner-written emotion vector, the QwenEmotion-generated emotion vector, and the waveform produced by the stock emotion route with planner-side duration control disabled.
Table~\ref{tab:stages} reports the agreement at each probe point independently.

The delivery phrase agrees with the declared direction on 84\% of Japanese items and 98\% of English items.
The planner-written emotion vector reaches 86\% and 89\%, respectively.
The two planner outputs agree with each other on 83\% of Japanese items and 89\% of English items.
Their disagreements differ across languages.
In Japanese, the delivery phrase alone is correct on 11 items and the planner-written vector alone on 14, with no significant asymmetry ($p=.69$).
In English, the delivery phrase alone is correct on 15 items, compared with one item for the vector alone ($p=.00052$).

The QwenEmotion-generated vector shows substantially lower agreement in Japanese than either planner representation.
Given the planner-generated delivery phrase, the QwenEmotion vector agrees with the declared direction on 46\% of Japanese items and 83\% of English items.
The Japanese rates are 61\% for arousal and 37\% for valence, while the corresponding English rates are 82\% and 84\%.
With the stock emotion route and planner-side duration control disabled, the final waveform is direction-correct on 55\% of Japanese items and 66\% of English items.
Because the probe points use representation-specific readouts, these rates should not be interpreted as conditional survival probabilities from one stage to the next.

\begin{table}[t]
\centering
\caption{Direction rate on the 150-item test sets, three plan seeds, per-item majority. Every condition receives the planner's duration decision unless marked w/o duration. Ar./Val.\ split the rate by declared axis. For the three treatments $p$ is a two-sided exact sign test against the base of the same language; for an ablation it is against the row it ablates.}
\label{tab:scoreboard}
\footnotesize
\setlength{\tabcolsep}{2pt}
\begin{tabular}{lcccc|cccc}
\toprule
& \multicolumn{4}{c|}{Japanese} & \multicolumn{4}{c}{English} \\
Condition & Dir. & Ar. & Val. & $p$ & Dir. & Ar. & Val. & $p$ \\
\midrule
Base: stock module & 65\% & 70\% & {62\%} & -- & 74\% & 77\% & 72\% & -- \\
Bypassed module & 68\% & 80\% & {61\%} & .58 & 79\% & 80\% & 79\% & .26 \\
Supervised module (SFT) & \textbf{75\%} & 77\% & {\textbf{74\%}} & .033 & 84\% & 91\% & 80\% & .036 \\
Learned interface & 66\% & \textbf{88\%} & 53\% & .60 & \textbf{89\%} & \textbf{95\%} & \textbf{85\%} & .008 \\
\midrule
\multicolumn{9}{l}{\emph{Ablations}} \\
Base w/o duration & 55\% & 59\% & {52\%} & .014 & 66\% & 66\% & 66\% & .008 \\
Base w/o conditioning & 61\% & 64\% & {60\%} & .61 & 74\% & 79\% & 71\% & 1.0 \\
Base w/o dur.\ \& cond. & 53\% & 54\% & {53\%} & .040 & 69\% & 64\% & 72\% & .39 \\
\bottomrule
\end{tabular}
\end{table}

\subsection{Can the interface be repaired?}
\label{ssec:res-repair}

Table~\ref{tab:scoreboard} compares three modifications of the emotion-control interface under the same planner-side duration control.

\textbf{Bypassing QwenEmotion.}
Passing the planner-written emotion vector and $\alpha$ directly to the prototype bank reaches 68\% in Japanese and 79\% in English.
Neither change is significant relative to Base ($p=.58$ and $.26$).
Thus, bypassing the stock mapping alone provides only modest evidence of improvement.

\textbf{Supervising QwenEmotion.}
We next retain the stock interface but explicitly supervise its text-to-vector mapping.
SFT raises the direction rate to 75\% in Japanese and 84\% in English, significantly above Base in both languages ($p=.033$ and $.036$).

\textbf{Learning a text-to-latent interface.}
Finally, we replace both QwenEmotion and the prototype bank with a learned text-to-latent mapping.
The learned route reaches 89\% overall in English ($p=.008$ against Base), the highest observed English rate, with 95\% arousal and 85\% valence.
In Japanese, its overall rate is 66\% ($p=.60$), but arousal reaches 88\%, the highest observed Japanese arousal rate.
The effect of the learned interface is therefore strongly axis- and language-dependent.

\textbf{Ablations.}
Removing planner-side duration control from Base reduces the direction rate from 65\% to 55\% in Japanese ($p=.014$) and from 74\% to 66\% in English ($p=.008$).
Removing planner-side emotion conditioning alone has a smaller effect, yielding 61\% in Japanese ($p=.61$) and 74\% in English ($p=1.0$).
Removing both duration and conditioning further lowers the Japanese rate to 53\% ($p=.040$), while the English rate becomes 69\% ($p=.39$).
These results indicate that planner-side duration control provides the clearest independent contribution to the full Base, while the effect of emotion conditioning is weaker in the aggregate metric and may depend on the acoustic axis.

\begin{table}[t]
\centering
\caption{Listening test on 30 random items per language, five judgments per item and condition. Judg.: share of correct judgments; Items: items (of 30) whose listener majority is correct, with a two-sided sign test against chance; Agr.: items (of 30) on which the listener majority agrees with the predictor's verdict.}
\label{tab:abx}
\footnotesize
\setlength{\tabcolsep}{2.5pt}
\begin{tabular}{lcccc|cccc}
\toprule
& \multicolumn{4}{c|}{Japanese} & \multicolumn{4}{c}{English} \\
Condition & Judg. & Items & $p$ & Agr. & Judg. & Items & $p$ & Agr. \\
\midrule
Base & 64\% & 19 & .20 & 20 & 62\% & 19 & .20 & 23 \\
Supervised (SFT) & 67\% & 20 & .10 & 21 & 69\% & 22 & .016 & 23 \\
Learned interface & 66\% & 20 & .10 & 20 & 63\% & 20 & .10 & 23 \\
\bottomrule
\end{tabular}
\end{table}

\subsection{Human validation}
\label{ssec:res-human}
Table~\ref{tab:abx} gives the result.
Judgment-level assignment accuracy ranges from 64--67\% in Japanese and 62--69\% in English.
At the item-majority level, the supervised module in English exceeds chance (22/30, $p=.016$), while the other conditions yield 19--20 correct items out of 30 ($p=.10$--$.20$).
Listener majorities agree with the predictor on 61/90 (68\%) Japanese and 69/90 (77\%) English item--condition pairs.
The test therefore provides descriptive evidence that the intended contrasts are perceptible and that the predictor often tracks listener judgments, but does not establish differences among the treatments, whose comparison rests on Table~\ref{tab:scoreboard}.

\section{Conclusion}
\label{sec:conclusion}
We introduced a bilingual persona--line conflict benchmark, and used it to probe the delivery phrase, emotion vector, and rendered speech while comparing bypassed, supervised, and learned emotion-control interfaces.
Our study is limited to one planner--synthesizer pipeline, two affective axes, and a reference-free acoustic scorer whose cross-lingual transfer remains imperfectly characterized.
Future work should move beyond coarse emotion categories and intermediate delivery descriptions by developing richer emotion captioners and evaluators, and by training models that reason directly from persona and context to continuous acting representations.

\vfill\pagebreak

\section{Acknowledgments}
\textbf{Ethics statement.} The listening study involved non-invasive perceptual judgments of speech. 
Participants were recruited through personal contacts, and were informed that their responses would be used for research and reported only in aggregate.  

\textbf{Use of Generative AI Disclosure.} Claude (Anthropic) was used to assist with language revision of the manuscript and experimental code development. All AI-generated outputs were reviewed and verified by the authors.

\bibliographystyle{IEEEbib}
\bibliography{refs}

\end{document}